\documentclass[11pt,a4paper]{article}
\usepackage[hyperref]{naaclhlt2018}
\usepackage{times}
\usepackage{latexsym}
\usepackage{amsmath}
\usepackage{amssymb}
\usepackage{url}

\usepackage{array}
\usepackage{graphicx}
\usepackage{hyperref}
\usepackage{xspace}

\usepackage{microtype}
\usepackage{todonotes}

\aclfinalcopy

\title{AMORE-UPF at SemEval-2018 Task 4: BiLSTM with Entity Library}

 \author{
 	Laura Aina\thanks{~~denotes equal contribution.}\hspace{4ex}Carina Silberer\footnotemark[1]\hspace{4ex}Ionut-Teodor Sorodoc\footnotemark[1]\vspace{1ex}\\ 
 	 \hspace{3ex}\textbf{Matthijs Westera\footnotemark[1]} \hspace{3ex}\textbf{Gemma Boleda}\vspace{2ex} \\
 	Universitat Pompeu Fabra \\
 	Barcelona, Spain \\
 	{\tt \{firstname.lastname\}@upf.edu}	
 }

\begin{document}
\newcommand{\fscore}{F$_1$-score\xspace}
	
\maketitle
\begin{abstract}
This paper describes our winning contribution to SemEval 2018 Task 4: \emph{Character Identification on Multiparty Dialogues}.
It is a simple, standard model with one key innovation, an \emph{entity library}.
Our results show that this innovation greatly facilitates the identification of infrequent characters.
Because of the generic nature of our model, this finding is potentially relevant to any task that requires effective learning from sparse or unbalanced data.
\end{abstract}

\section{Introduction} \label{sec:introduction}


SemEval 2018 Task 4 is an \textbf{entity linking} task on multiparty
dialogue.%
\footnote{\url{https://competitions.codalab.org/competitions/17310}}
It consists in predicting the referents of nominals that refer to a person, such as \textit{she}, \textit{mom}, \textit{Judy} -- henceforth \emph{mentions}.
The set of possible referents is given beforehand, as well as the set of mentions to resolve.
The dataset used in this task is based on \citet{ChenChoi:16} and \citet{ChenZhouChoi:17}, and consists of dialogue from the TV show \emph{Friends} in textual form.

Our main interest is whether deep learning models for tasks like entity linking can benefit from having an explicit \textbf{entity library}, i.e., a component of the neural network  that stores entity representations learned during training.
To that end, we add such a component to an otherwise relatively basic model -- a bidirectional LSTM (long short-term memory; \citealt{hochreiter1997long}), the standard neural network model for sequential data like language.
Training and evaluating this model on the task shows that the entity
library is beneficial in the case of infrequent entities.%
\footnote{Source code for our model and for the training procedure is published on \url{https://github.com/amore-upf/semeval2018-task4}.}

\section{Related Work}
\label{sec:related_work}

Previous entity linking tasks concentrate on linking mentions to Wikipedia pages (\citealt{bunescu2006using, mihalcea2007wikify} and much subsequent work; for a recent approach see \citealt{francis2016capturing}).
By contrast, in the present task (based on \citealt{ChenChoi:16, ChenZhouChoi:17}) only a list of entities is given, without any associated encyclopedic entries.
This makes the task more similar to the way in which a human audience might watch the TV show, in that they are initially unfamiliar with the characters.
What also sets the present task apart from most previous tasks is its focus on multiparty dialogue (as opposed to, typically, newswire articles).

A task that is closely related to entity linking is \emph{coreference resolution}, i.e., the task of clustering mentions that refer to the same entity (e.g., the CoNLL shared task of \citealt{pradhan2011conll}).
Since mention clusters essentially correspond to entities (an insight central to the approaches to coreference in \citealt{haghighi2010coreference,clark2016improving}), the present task can be regarded as a type of coreference resolution, but one where the set of referents to choose from is given beforehand.

Since our main aim is to test the benefits of having an entity library, in other respects our model is kept more basic than existing work both on entity linking and on coreference resolution (e.g., the aforementioned approaches, as well as \citealt{wiseman2016learning}; \hbox{\citealt{lee2017end}}, \hbox{\citealt{francis2016capturing}}).
For instance, we avoid feature engineering, focusing instead on the model's ability to learn meaningful entity representations from the dialogue itself.
Moreover, we deviate from the common strategy to entity linking of incorporating a specialized coreference resolution module (e.g., \citealt{ChenZhouChoi:17}).

\section{Model Description} \label{sec:model_description}

\begin{figure}[t] 
	\centering
	\includegraphics[width=.85\columnwidth]{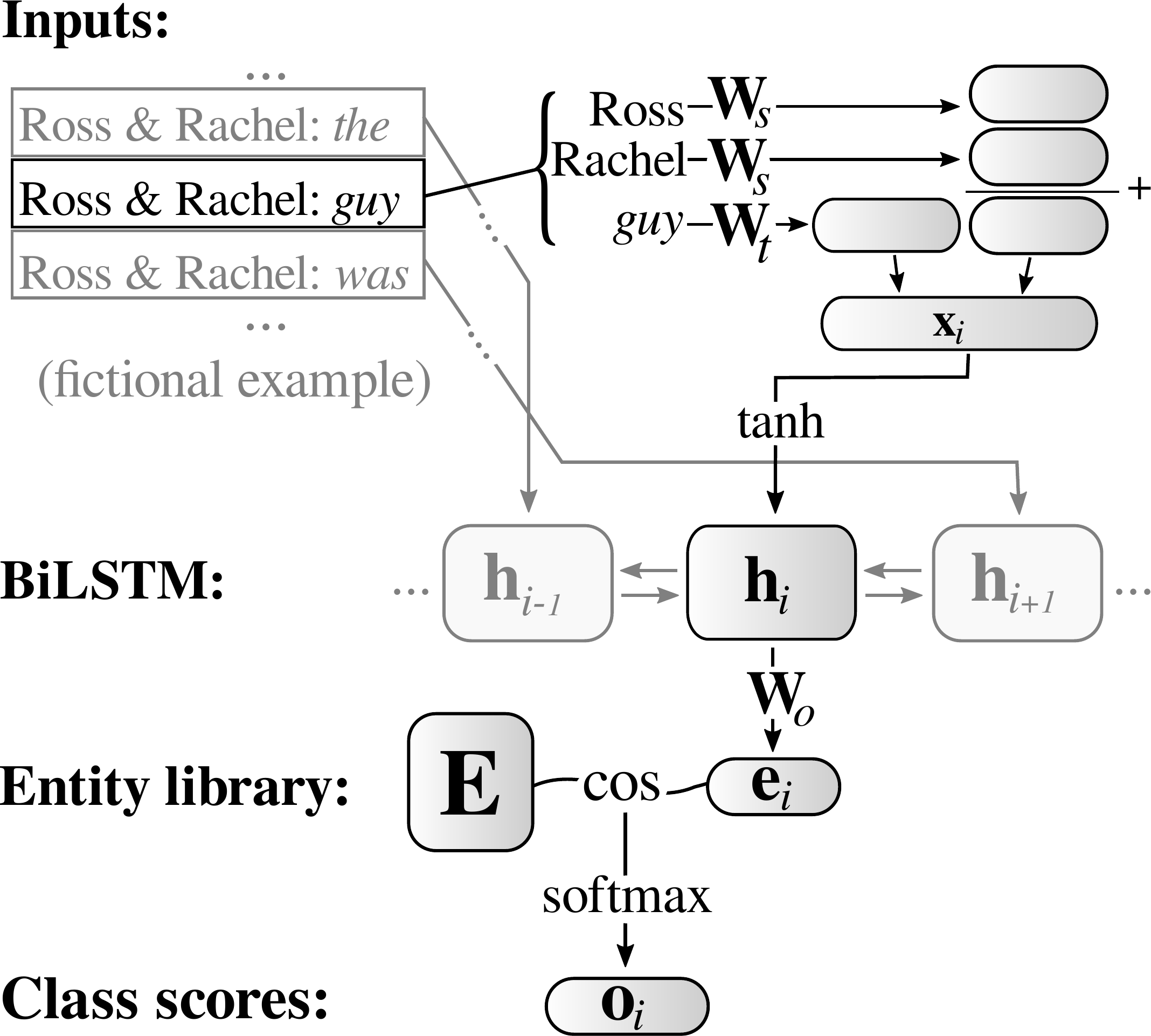}
	\caption{The AMORE-UPF model (bias not depicted).}
	\label{fig:modeldiagram}
	\vspace{-.2cm}
\end{figure}

We approach the task of character identification as one of multi-class classification. Our model is depicted in Figure~\ref{fig:modeldiagram}, with inputs in the top left and outputs at the bottom.
In a nutshell, our model is a bidirectional LSTM (long short-term memory, \citealt{hochreiter1997long}) that processes the dialogue text and 
resolves mentions, through a comparison between the LSTM's hidden state (for each mention) to vectors in a learned entity library.\looseness=-1

The model is given chunks of dialogue, which it processes token by token.
The $i$\textsuperscript{th} token $\mathbf{t}_i$ and its speakers $S_i$ (typically a singleton set) are represented as one-hot vectors, embedded via two distinct embedding matrices ($\mathbf{W}_t$ and $\mathbf{W}_s$, respectively) and finally concatenated to form a vector~$\mathbf{x}_i$ 
(Eq.~\ref{eq:input}; see also 
Figure~\ref{fig:modeldiagram}).
In case $\mathbf{S}_i$ contains multiple speakers, their embeddings are summed.
\begin{equation}
	\label{eq:input} \mathbf{x}_i = \mathbf{W}_t \ \mathbf{t}_i \ \| \displaystyle \sum_{\mathbf{s} \in S_i} \mathbf{W}_s \ \mathbf{s}
\end{equation}
We apply an activation function~\mbox{$f$ $( = \text{tanh})$}. 
The hidden state $\overrightarrow{\mathbf{h}}_{\! i}$ of a \emph{uni}directional LSTM for the $i$\textsuperscript{th} input is recursively defined as a combination of that input with the LSTM's previous hidden state $\overrightarrow{\mathbf{h}}_{\! i-1}$.
For a \emph{bi}directional LSTM, the hidden state $\mathbf{h}_i$ is a concatenation of the hidden states $\overrightarrow{\mathbf{h}}_{\! i}$ and $\overleftarrow{\mathbf{h}}_{\! i}$ of two unidirectional LSTMs which process the data in opposite directions (Eq.~\ref{eq:hidden}; see also Figure~\ref{fig:modeldiagram}).
In principle, this enables a bidirectional LSTM to represent the entire dialogue with a focus on the current input, including for instance its relevant dependencies on the context.\looseness=-1
\begin{equation}
	\label{eq:hidden} \mathbf{h}_i = \text{BiLSTM}(f(\mathbf{x}_i), \overrightarrow{\mathbf{h}}_{\! i-1}, \overleftarrow{\mathbf{h}}_{\! i+1})
\end{equation}

In the model, learned representations of each entity are stored in the \textit{entity library} $\mathbf{E} \in \mathbb{R}^{N \times k}$ (see Figure~\ref{fig:modeldiagram}): 
$\mathbf{E}$ is a matrix which represents each of $N$~entities through a  $k$-dimensional vector, and whose values are updated (only) during training.
For every token $\mathbf{t}_i$ that is tagged as a mention,%
\footnote{For multi-word mentions this is done only for the last token in the mention.}
we map the corresponding hidden state~$\mathbf{h}_i$ to a vector $\mathbf{e}_i \in \mathbb{R}^{1 \times k}$. 
This extracted representation is used to retrieve the (candidate) referent of the  mention from the entity library: 
The similarity of~$\mathbf{e}_i$ to each entity representation stored in $\mathbf{E}$ is computed using cosine, and softmax is then applied to the resulting similarity profile to obtain a probability distribution \mbox{$\mathbf{o}_i \in [0,1]^{1 \times N}$} over entities (`class scores' in Figure~\ref{fig:modeldiagram}):\looseness=-1
\begin{equation}
	\label{eq:output1} \mathbf{o}_i =  \text{softmax} ( \text{cosine} (\mathbf{E}, \underbrace{(\mathbf{W}_o \ \mathbf{h}_i + \mathbf{b})}_{\mathbf{e}_i})
\end{equation} 
At testing time, the model's prediction $\hat{c}_i$ for the $i$\textsuperscript{th} token is the entity with highest probability:\looseness=-1
\begin{equation}
	\label{eq:argmax} \hat{c}_i = \text{argmax} (\mathbf{o}_i)
\end{equation}
We train the model with backpropagation, using negative log-likelihood as loss function. Besides the BiLSTM parameters, we optimize $\mathbf{W}_t$, $\mathbf{W}_s$, $\mathbf{W}_o$, $\mathbf{E}$ and $\mathbf{b}$.
We refer to this model as \textbf{AMORE-UPF}, our team name in the SemEval competition. 
Note that, in order for this architecture to be successful, $\mathbf{e}_i$ needs to be as similar as possible to the entity vector of the entity to which mention $t_i$ refers. 
Indeed, the mapping $\mathbf{W}_o$ should effectively specialize in ``extracting'' entity representations from the hidden state because of the way its output $\mathbf{e}_i$ is used in the model---to do entity retrieval.
Our entity retrieval mechanism is inspired by the attention mechanism of~\citet{bahdanau2016neural}, that has been used in previous work to interact with an external memory~\cite{Sukhbaatar2015,boleda+iwcs17}.
To assess the contribution of the entity library, we compare our model to a similar architecture which does not include it (\textbf{NoEntLib}).
This model 
 directly applies softmax to a linear mapping of the hidden state (Eq.~\ref{eq:output2}, replacing Eq.~\ref{eq:output1} above).
\begin{equation}
	\label{eq:output2}\mathbf{o}_i =   \text{softmax} (\mathbf{W}_o \ \mathbf{h}_i + \mathbf{b})
\end{equation}

\section{Experimental Setup}
\label{sec:experiments}

\paragraph{Data}
We use the training and test data provided for SemEval~2018 Task~4, which span the first two seasons of the TV show \textsl{Friends}, divided into scenes (train: $374$~scenes from $47$~episodes; test: $74$ scenes from $40$ episodes).%
\footnote{The organizers also provided data divided by episodes rather than scenes, which we didn't use.} 
In total, the training and test data contain~13,280 and 2,429~nominal mentions (e.g.,~\textsl{Ross, I}; Figure~\ref{fig:example}), respectively, which are annotated with the ID of the entity to which they refer (e.g.,~335, 183).
The utterances are further annotated with the name of the speaker (e.g.,~\textsc{Joey Tribbiani}). 
Overall there are 372~entities in the training data (test data:~106).
\begin{figure}[t]
	\fbox{
		\parbox{.45\textwidth}{
			\textsc{Joey Tribbiani} (183):\\
			\begin{tabular}{@{~}r@{~}r@{~}r@{~}r@{~}r@{~}lr}
				"\ldots \textsl{see} 	& \textsl{\underline{Ross},} 
				& \textsl{because \underline{I}} 
				& \textsl{think \underline{you}} 
				& \textsl{love}  \textsl{\underline{her}} 
				& ." \\
				& 335 
				& 183 
				& 335
				& 306
				& 
			\end{tabular}}}
			\vspace{-1ex}
			\caption{Example of the data provided for the SemEval~2018 Task~4. It shows the speaker (first line) of the utterance (second line) and the ids of the entities to which the target mentions (underlined) refer (last line). \label{fig:example}}
			\vspace{-.7cm}
		\end{figure}
Our models do not use any of the provided automatic linguistic annotations, such as PoS or named entity tags.

We additionally used the publicly available 300-dimensional word vectors that were pre-trained on a Google News corpus with the word2vec Skip-gram model \citep{mikolov2013distributed}.%
\footnote{The word vectors are available at \url{https://code.google.com/archive/p/word2vec/}.}

\paragraph{Parameter settings}
Using 5-fold cross-validation on the training data, we performed a random search over the hyperparameters and chose those which yielded the best mean F1-score.
Specifically, our submitted model is trained in batch mode using the Adam optimizer \cite{kingma2017adam} with a learning rate of~$0.0005$.
Each batch covers $24$ scenes, which are given to the model in chunks of $757$~tokens.
The token embeddings ($\mathbf{W}_t$) are initialized with the word2vec vectors. 
Dropout rates of $0.008$ and $0.0013$ are applied on the input~$\mathbf{x}_i$ and hidden layer~$\mathbf{h}_i$ of the LSTM, respectively. 
The size of~$\mathbf{h}_i$ is set to $459$~units, 
the embeddings of the entity library~$\mathbf{E}$ and speakers~$\mathbf{W}_s$ are set to $k~\mbox{= 134}$~dimensions.\looseness=-1

Other configurations, including randomly initialized token embeddings, 
weight sharing between $\mathbf{E}$ and $\mathbf{W}_s$,
self-attention \cite{bahdanau2016neural} on the input layer, a uni-directional LSTM, and rectifier or linear activation function~$f$ on the input embeddings did not improve performance.

For the final submission of the answers for the test data, we created an ensemble model by averaging the output (Eq.~\ref{eq:output1}) of the five models trained on the different folds.

\section{Results}
\label{sec:results}
\begin{table}[t]
	\centering
	\begin{tabular}{|l@{~}|c@{~}|c@{~}|c|c|}
		\hline
		& \multicolumn{2}{c|}{all entities} &
		\multicolumn{2}{c|}{main entities} \\
		Models & F$_1$ & Acc & F$_1$ & Acc \\
		\hline \hline
		AMORE-UPF & $\mathbf{41.1^{**}}$ & $\mathbf{74.7^{**}}$ & $79.4$ & $77.2$\\
		NoEntLib  & $26.4$ & $71.6$ & $\mathbf{79.5}$ & $\mathbf{77.5}$\\
		\hline
	\end{tabular}
	\caption{Results obtained for the submitted AMORE-UPF model and a  variant of it that does not use an entity library (NoEntLib). Best results are in boldface. Differences with respect to the 2nd row marked by `**' are significant at the $0.001$~probability level (see text). \label{tab:main_results}}
\end{table}
Two evaluation conditions were defined by the organizers -- \emph{all entities} and \emph{main entities} -- with macro-average \fscore and label accuracy as the official metrics, and macro-average \fscore in the \emph{all entities} condition applied to the leaderboard.
The \emph{all entities} evaluation has 67 classes: 66 for entities that are mentioned at least 3~times in the test set and one grouping all others. 
The \emph{main entities} evaluation has 7 classes, 6 for the main characters and one for all the others. 
Among all four participating systems in this SemEval task our model achieved the highest score on the \emph{all entities} evaluation, and second-highest on the \emph{main entities} evaluation.\looseness=-1

\begin{figure}[t]
	\includegraphics[scale=.55]{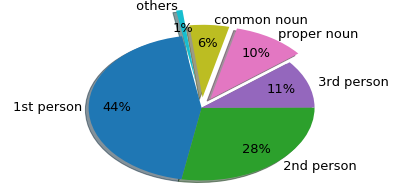}
	\vspace{-4ex}
	\caption{Distribution of all $2,429$~target mentions in the test data in terms of their part-of-speech.\label{fig:distribution}}
	\vspace{-.3cm}
\end{figure}

Table~\ref{tab:main_results} gives our results in the two evaluations, comparing the models described in Section~\ref{sec:experiments}.
While both models perform on a par on main entities, AMORE-UPF outperforms NoEntLib by a substantial margin when all characters are to be predicted 
(+15 points in \fscore, +3 points in accuracy; Table~\ref{tab:main_results}).%
\footnote{The mean difference between the single models (trained on a single fold) and the ensemble AMORE-UPF is between $\mbox{-1.3}$~points (accuracy main entities, std~\mbox{$=1.3$)} and~$\mbox{-4.2}$~points (\fscore all entities, std~\mbox{$=1.3$)}.}
The difference between the models with/without an entity library are statistically significant based on approximate randomization tests \cite{noreen1989computer}, with the significance level~\mbox{$p < 0.001$}. 
This shows that the use of an entity library can be beneficial for the linking of rarely mentioned characters.\looseness=-1

\begin{figure}[t]
	\includegraphics[scale=.53]{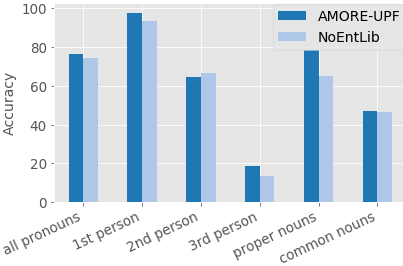}
	\vspace{-2ex}
	\caption{Aaccuracy of the models on \textsl{all entities} depending on the part-of-speech of the target mentions.\label{fig:pronouns_acc}}
	\vspace{-.4cm}
\end{figure}

Figure~\ref{fig:distribution} shows that most of the target mentions in the test data fall into one of five grammatical categories. 
The dataset contains mostly pronouns (83\%), with a very high percentage of first person pronouns (44\%). 
Figures~\ref{fig:pronouns_f1} and~\ref{fig:pronouns_acc} present the accuracy and \fscore which the two models described above obtain on \textsl{all entities} for different categories of mentions. 
The entity library is beneficial when the mention is a first person pronoun or a proper noun (with an increase of~30 points in \fscore for both categories;  Figure~\ref{fig:pronouns_f1}), and closer inspection revealed that this effect was larger for rare entities.\looseness=-1

\section{Discussion}
\label{sec:conclusion}

The AMORE-UPF model consists of a bidirectional LSTM linked to an entity library. 
Compared to an LSTM without entity library, NoEntLib, the AMORE-UPF model performs particularly well on rare entities, which explains its top score in the \emph{all entities} condition of SemEval 2018 Task 4.
This finding is encouraging, since rare entities are especially challenging for the usual approaches in NLP, due to the scarcity of information about them.
\begin{figure}[t]
	\includegraphics[scale=.53]{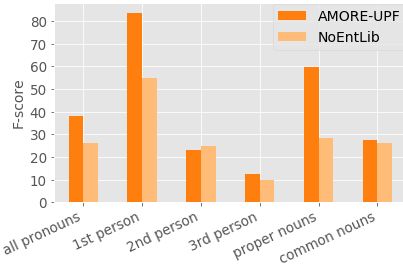}
	\vspace{-2ex}
	\caption{\fscore of the models on \textsl{all entities} depending on the part-of-speech of the target mentions.\label{fig:pronouns_f1}}
	\vspace{-.4cm}
\end{figure}

We offer the following explanation for this beneficial effect of the entity library, as a hypothesis for future work.
Having an entity library requires the LSTM of our model to output some representation of the mentioned entity, as opposed to outputting class scores more or less directly as in the variant NoEntLib.
Outputting a meaningful entity representation is particularly easy in the case of first person pronouns and nominal mentions (where the effect of the entity library appears to reside; Figure~\ref{fig:pronouns_f1}): 
the LSTM can learn to simply forward the speaker embedding unchanged in the case of pronoun \textit{I}, and the token embedding in the case of nominal mentions.
This strategy does not discriminate between frequent and rare entities; it works for both alike.
We leave further analyses required to test this potential explanation for future work.

Future work may also reveal to what extent the induced entity representations  
may be useful in others, to what extent they encode entities' attributes and relations (cf. \citealt{gupta2015distributional}), and to what extent a module like our entity library can be employed elsewhere, in natural language processing and beyond.

\section*{Acknowledgments}
This project has received funding from the European Research Council (ERC) under the European Union’s Horizon 2020 research and innovation programme (grant agreement No 715154), and from the Spanish Ram\'on y Cajal programme (grant RYC-2015-18907). We are grateful to the NVIDIA Corporation for the donation of GPUs used for this research. We are also very grateful to the Pytorch developers. This paper reflects the authors' view only, and the EU is not responsible for any use that may be made of the information it contains.
\begin{flushright}
\includegraphics[width=0.8cm]{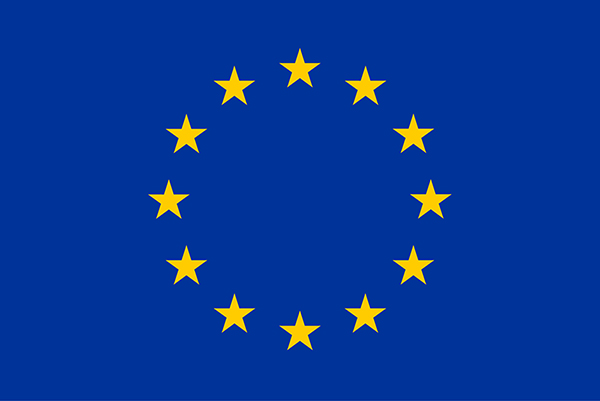}  
\includegraphics[width=0.8cm]{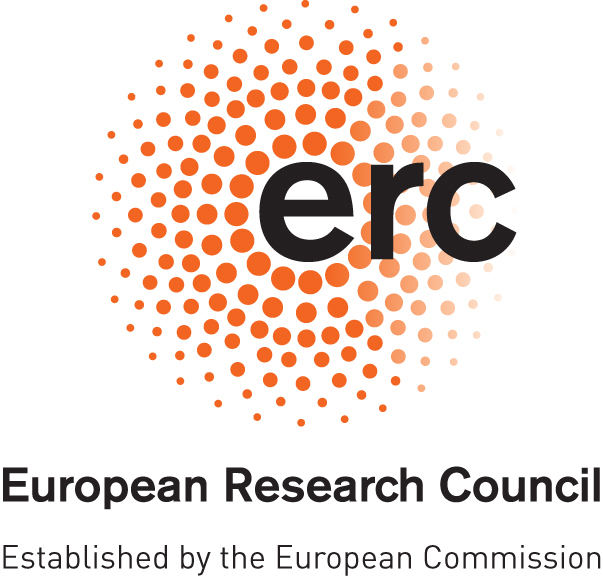} 
\end{flushright}

\bibliography{semeval2018-amore_upf}
\bibliographystyle{acl_natbib}

\end{document}